\pdfoutput=1
\documentclass[11pt]{article}

\usepackage[]{acl2023}

\usepackage{times}
\usepackage{latexsym}

\usepackage[T1]{fontenc}
\usepackage[utf8]{inputenc}

\usepackage{microtype}
\usepackage{graphicx}
\usepackage{amsmath}
\usepackage{tabularx}

\usepackage{tikz}
\usepackage{xcolor}

\definecolor{datapillcolor}{RGB}{7, 55, 99}
\definecolor{modelpillcolor}{RGB}{250, 174, 0}
\definecolor{losspillcolor}{RGB}{255, 255, 255}
\definecolor{transformcolor}{RGB}{63, 141, 87}
\definecolor{october}{RGB}{63, 141, 87}

\renewcommand\fbox{\fcolorbox{black}{white}}

\newcommand{\datapill}{\colorbox{datapillcolor}{\textcolor{white}{\textsc{\textbf{\tiny Data}}}}}
\newcommand{\modelpill}{\colorbox{modelpillcolor}{\textcolor{white}{\textsc{\textbf{\tiny Model}}}}}
\newcommand{\transformpill}{\colorbox{transformcolor}{\textcolor{white}{\textsc{\textbf{\tiny Transform}}}}}
\newcommand{\losspill}{\fbox{\textsc{\textbf{\tiny loss}}}}
\newcommand{\declarelogo}[0]{\includegraphics[height=.02\textwidth]{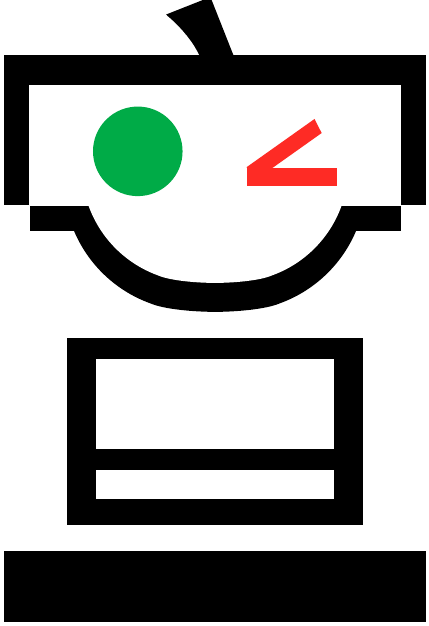}}
\newcommand{\abhinav}[1]{\textcolor{black}{#1}}

\usepackage{inconsolata}

\usepackage{cleveref}
\crefname{section}{§}{§§}
\Crefname{section}{§}{§§}
\usepackage{enumitem}

\title{A Comprehensive Survey of Sentence Representations:\\From the \textsc{BERT} Epoch to the \textsc{ChatGPT} Era and Beyond}
\author{Abhinav Ramesh Kashyap$^{\alpha}$, 
Thanh-Tung Nguyen$^{\alpha}$, Viktor Schlegel$^{\alpha}$, \\ \textbf{Stefan Winkler$^{\alpha\beta}$, See-Kiong Ng$^{\beta}$,  Soujanya Poria$^{\declarelogo}$} \\
${\alpha}$: ASUS Intelligent Cloud Services (AICS), Singapore\\
${\beta}$: National University of Singapore, Singapore\\
${\declarelogo}$: DeCLaRe Lab, Singapore University of Technology and Design, Singapore \\
\fontsize{10}{10}\texttt{\{abhinav\_kashyap,thomas\_nguyen,viktor\_schlegel,stefan\_winkler\}@asus.com}\\
\fontsize{10}{10}\texttt{seekiong@nus.edu.sg, sporia@sutd.edu.sg} 
}

\begin{document}
\maketitle
\begin{abstract}
Sentence representations are a critical component in NLP applications such as retrieval, question answering, and text classification. They capture the meaning of a sentence, enabling machines to understand and reason over human language. In recent years, significant progress has been made in developing methods for learning sentence representations, including unsupervised, supervised, and transfer learning approaches. \abhinav{However there is no literature review on sentence representations till now}. In this paper, we provide an overview of the different methods for sentence representation learning, focusing mostly on  deep learning models. We provide a systematic organization of the literature, highlighting the key contributions and challenges in this area. Overall, our review highlights  the importance of this area in natural language processing, the progress made in sentence representation learning, and the challenges that remain. We conclude with directions for future research, suggesting potential avenues for improving the quality and efficiency of sentence representations.
\end{abstract}

\begin{figure*}
    \centering
    \includegraphics[width=\textwidth]{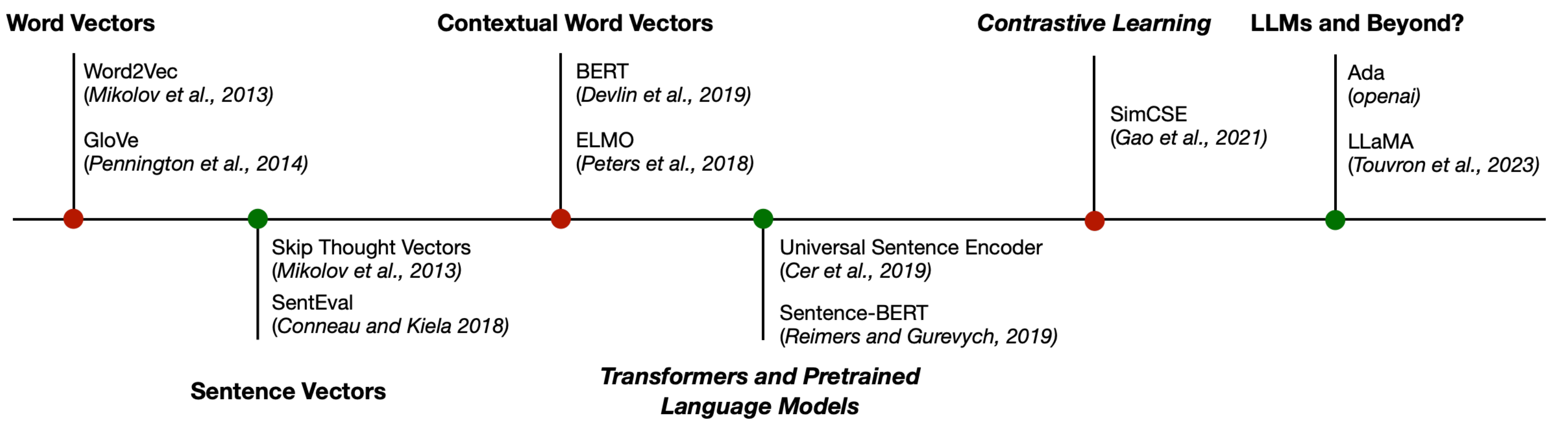}
    \caption{Illustration of some of the milestones in Sentence Representation Learning Research}
    \label{fig:timeline}
\end{figure*}

\section{Introduction}

%Sentence representation 
The {\it sentence}, together with the {\it word}, are the two fundamental grammatical units of human language. Representing sentences for machine learning, which involves transforming a sentence into a vector or a fixed-length representation, is an important component of NLP. The quality of these  representations affects the performance of downstream NLP tasks like text classification and text similarity \cite{conneau-kiela-2018-senteval}.

Deep neural networks have played a major role in obtaining sentence representations. While there have been significant advancements in the development of large language models (LLMs) such as GPT-3 \cite{brown-fewshot},  BLOOM \cite{workshop2023bloom}, they learn through effective word representations and modelling of the language at the (next) word level. Endowing models with the ability to learn effective representations of  higher linguistic units beyond words -- such as sentences -- is useful.

For instance, sentence representations can help in  retrieving semantically similar documents prior to generation. LangChain\footnote{https://github.com/hwchase17/langchain} and various other frameworks like DSPy \cite{khattab2023dspy}, have underscored the critical demand for proficient sentence representations. The documents retrieved serve as valuable resources for generating fact-based responses, using custom documents to address user queries, and fulfilling  other essential functions. 

However, current language models  exhibit drawbacks in obtaining sentence representations out-of-the-box. For instance, \citet{ethayarajh-2019-contextual} showed that out-of-the-box representations from BERT \cite{devlin-etal-2019-bert} are fraught with problems such as anisotropy---representations occupying a narrow cone, making every representation closer to all others. Also, they are impractical for real-life applications: finding the best match for a query takes hours \cite{reimers-gurevych-2019-sentence}.

To overcome the inadequacy of directly using sentence representations from language models, numerous methods have been developed. Several works have proposed to post-process the representations from BERT to alleviate the anisotropy \citep{li-etal-2020-sentence, huang-etal-2021-whiteningbert-easy} or repurpose representations from different layers of the model \cite{kim-etal-2021-self}. But there has been a steadily growing body of works that move away from such post-processing and introduce new methods.% \cite{chen-etal-2022-generate, ni-etal-2022-sentence}

Perhaps due to the rapid advancements in the field (\Cref{fig:timeline}), there are no literature reviews discussing the diverse range of techniques for learning sentence representations.  The present paper offers a review of these techniques, with a specific emphasis on deep learning methods. Our review caters to two audiences: (a) Researchers from various fields seeking to get insights into recent breakthroughs in sentence representations, and (b) researchers aiming to advance the field of sentence representations.

\subsection{Overview}
We structure our literature review as follows: % (cf.\ \Cref{fig:overview}): 
\begin{itemize}[noitemsep, leftmargin=*,labelindent=0em,itemindent=0em]
\item \Cref{sec:background} provides a brief history of methods to learn sentence representations and the different components of a modern framework.

\item \Cref{sec:supervised} provides a review of supervised sentence representations that use labeled data to learn sentence representations.

\item \Cref{sec:unsupervised} reviews methods that use unlabeled data to learn sentence representations (also called unsupervised sentence representation learning), a major focus of recent methods. 

\item \Cref{sec:other} describes methods that draw inspiration from other fields such as computer vision. 

\item \Cref{sec:analysis} provides a discussion of trends and analysis. 

\item \Cref{sec:challenges} discusses the challenges  and suggests some future directions for research.
\end{itemize}

\section{Background}
\label{sec:background}

% \subsection{Word Representations}
% A well-known challenge in NLP is creating continuous dense vector representations of words in high-dimensional spaces to capture their semantic and syntactic meaning. The most widely used algorithm for creating word embeddings is Word2Vec~\cite{word2vec-mikolov}.

% Traditional approaches to representing words before Word2Vec, like one-hot encoding or bag-of-words, have a number of drawbacks: They require a lot of memory to hold sparse vectors and fail to capture the links between words or their meaning.

% By using a neural network to learn word embeddings, Word2Vec solved these issues. The model trains neural networks using a large corpus of text as input to predict the likelihood of a word given its context or vice versa. The weights of the network are changed during training to reduce the discrepancy between the expected and actual probabilities. The network weights are employed as the word embeddings after training is finished. It has been widely used and inspired other models such as GloVe \cite{pennington-etal-2014-glove} and fastText \cite{joulin2016bag}.

\begin{figure}[t!]
    \centering
    \includegraphics{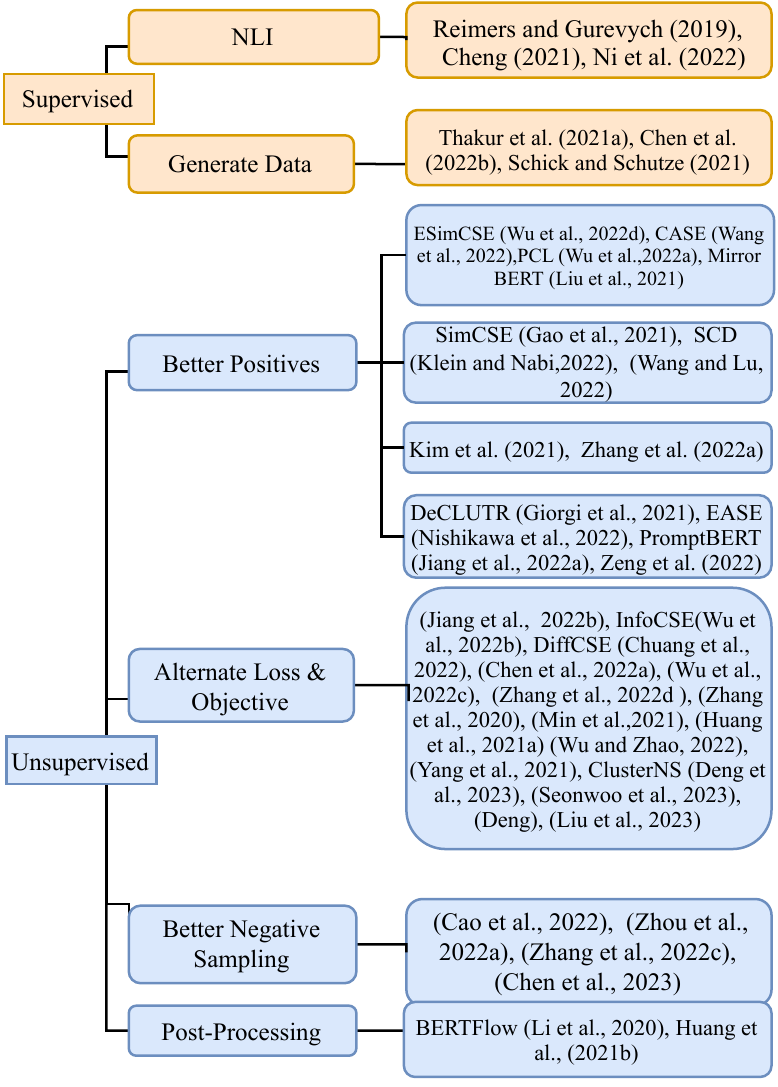}
    \caption{Overview of sentence representation methods.}
    \label{fig:overview}
\end{figure}

\subsection{Sentence Representations}
Before the advent of neural networks, bag-of-words models were commonly used to represent sentences, but they suffered from limitations such as being unable to capture the relationships between words or the overall structure of the sentence. 

Numerous efforts have aimed to improve sentence representations (\Cref{fig:timeline}). Inspired by Word2Vec \cite{word2vec-mikolov, pennington-etal-2014-glove}, \citet{skipthought-kiros} trained neural networks to predict the surrounding sentences of a given target sentence. Subsequently, \citet{conneau-kiela-2018-senteval} employed various recurrent neural networks (RNNs) to produce sentence embeddings, exploring their linguistic attributes, including part-of-speech tags, verb tense and named entity recognition. Notably, this study utilized natural language inference (NLI) data for neural network training, predating the emergence of extensive pretrained models such as BERT \cite{devlin-etal-2019-bert}. BERT and similar models have since served as a foundation for enhancing sentence representations. Exploring whether Large Language Models will ignite advancements in sentence representations or if pretrained language models like BERT remain pivotal is a crucial inquiry within today's context. (\Cref{sec:analysis})

\subsection{Components of Sentence Representations}
\label{subsec:sentence-representation-components}

\begin{figure}[t!]
    \centering
    \includegraphics[width=0.5\columnwidth]{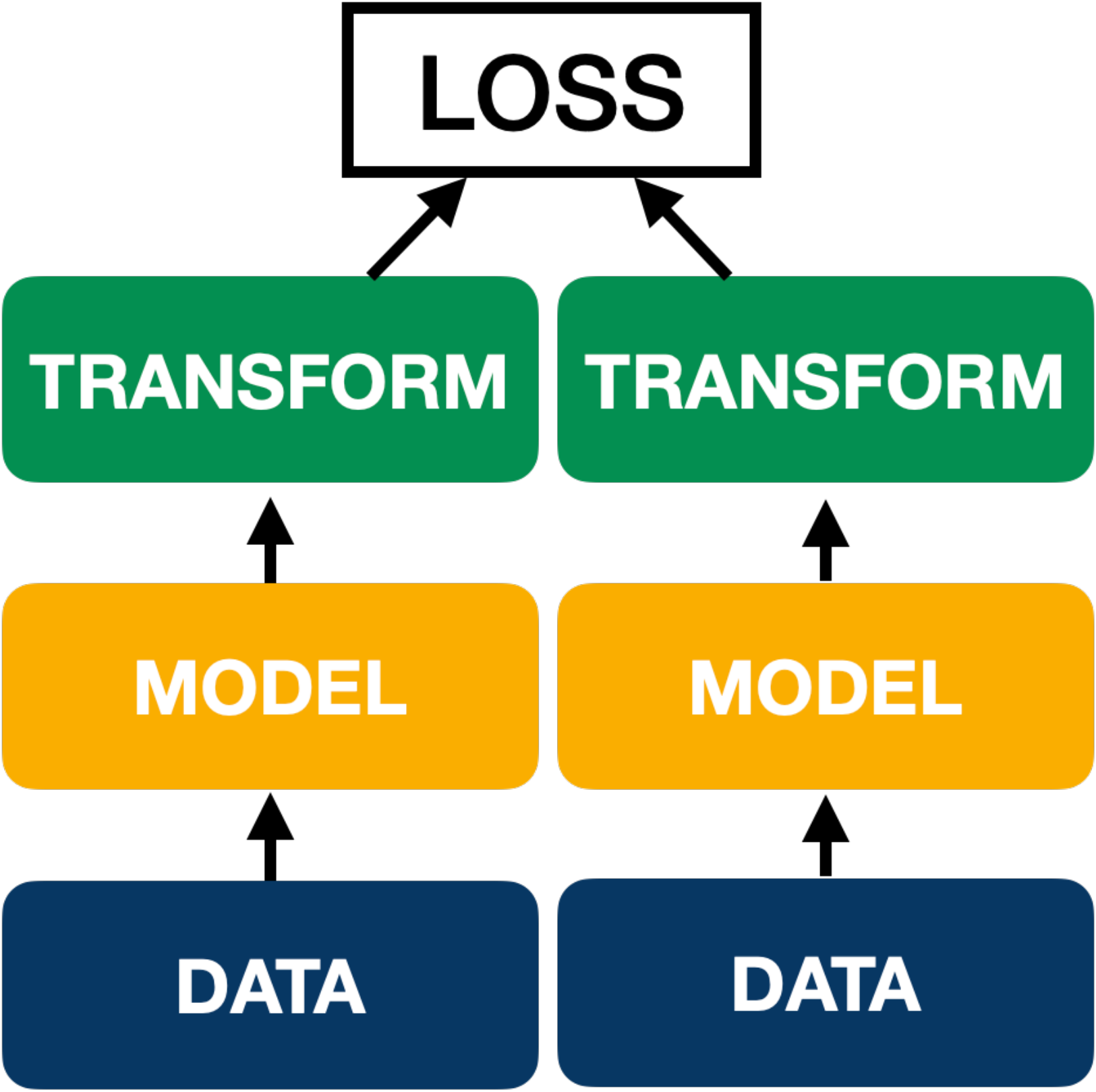}
    \caption{The components of an architecture to learn sentence representations. There are four main components: 1) \textbf{Data} - Obtaining positive and negative examples either using supervised data or some transformation 2) \textbf{Model} - Generally a pretrained model that has been trained on large quantities of gneeral text. 3) \textbf{Transform} - Some transformation applied to the representations from the model to obtain sentence representations, and 4) \textbf{Loss} - Losses that bring semantically similar sentences closer together and others apart. }
    \label{fig:components}
\end{figure}

Neural networks have become the de-facto standard for learning sentence representations. The network takes two sentences as input and creates a vector for each sentence. These vectors are then trained to be similar for sentences that mean the same thing and different for sentences with different meanings. Learning sentence representations using neural networks involves the following generic components (\Cref{fig:components}):
\begin{enumerate}[noitemsep, leftmargin=*,labelindent=0em,itemindent=0em]
    \item \textbf{Data}: Data used for learning sentence representations consists of pairs of semantically similar sentences, which can be either annotated by humans or generated through transformations to create positive and negative sentence pairs. (cf.\ \Cref{subsec:better-positives,subsec:better-negative-sampling}).
    \item \textbf{Model}: A sentence representation extraction model is a neural network backbone model unless specified otherwise. The backbone model can take the form of an RNN or a pretrained transformer model like BERT \cite{devlin-etal-2019-bert} or T5 \cite{t5-raffel}.
    \item \textbf{Transform}: Neural network representations are not well suited for use as sentence representations directly. While the \texttt{[CLS]} representations from BERT can serve as such, \citet{reimers-gurevych-2019-sentence} propose a pooling mechanism to obtain sentence representations by aggregating the token representations. The transformation required depends on the  model type. 
    \item \textbf{Loss}: Contrastive learning is often used for sentence representations. The objective is to bring semantically similar examples closer together while pushing dissimilar examples further apart. Specifically, given a set of example pairs $\mathcal{D} = \{x_i, x_i^{p}\}$, a model is used to obtain representations for each pair, denoted $h_i$ and $h_i^p$. The contrastive loss for an example is:
    \begin{equation*}
        l_i = -\log \frac{e ^{sim(h_i, h_i^p)}}{\sum_{j=1}^N e^{sim(h_i, h_j)}}
    \end{equation*}
        where N is the size of a mini-batch,  $sim(\cdot,\cdot)$ is the similarity function which plays a crucial role. However, when selecting an appropriate loss function, several factors need to be considered. These factors include the choice of similarity measures and the characteristics of the negative examples.

\end{enumerate}
    \abhinav{The different components have disproportionate effects in learning sentence representations. While \textbf{Model} has played an important role and has brought the most advances in learning sentence representations, \textbf{Data} cannot be disregarded. Most of the innovations have been concentrated in obtaining the right data for training.}
      
In their influential paper, \citet{reimers-gurevych-2019-sentence} utilized this versatile framework to generate highly effective sentence embeddings, which has subsequently served as a cornerstone for further research. This framework, commonly referred to as the bi-encoder or \abhinav{Siamese network} approach, involves encoding the \textit{query} and \textit{candidate} separately. \abhinav{This does not encourage interactions betweeen words. Encouraging word interactions can be achieved through a cross encoder, where the query and candidate are concatenated and encoded by a single model. However, this approach is computationally expensive and we have omitted it in this paper. In contrast, the 
Siamese BERT network pre-computes query and candidate vectors, enabling fast retrieval.}

\Cref{fig:overview} illustrates the progression of work aimed at improving sentence representations. Two primary approaches stand out: supervised and unsupervised methods. For a clearer understanding of innovations, we categorize these methods based on variations of common techniques. Each category identifies contributions that target specific components (\Cref{fig:components}): The \textit{Better Positives} category focuses on refining augmentation techniques, primarily addressing the \textbf{Data} component. Conversely, the \textit{Alternate Loss and Objectives} category explores improvements in the contrastive \textbf{Loss} function. These dynamic interactions between categories are further depicted in \Cref{tab: results}.

\section{Supervised Sentence Representations}
\label{sec:supervised}
Natural language understanding involves intricate reasoning. One way to learn better sentence representations is by excelling at tasks that demand reasoning. Large-scale supervised datasets for natural language understanding have emerged over the years: SNLI \cite{bowman-etal-2015-large}, MNLI \cite{williams-etal-2018-broad}, ANLI \cite{nie-etal-2020-adversarial}. To that end, neural network  methods  utilize supervised datasets to learn sentence representations.

\renewcommand{\arraystretch}{1.1}
\begin{table*}[t]
    \centering 
    \footnotesize 
    \begin{tabular}{|p{4cm}|c|c|p{2cm}|p{2cm}|c|}
    \hline
    \textsc{\textbf{Name}} & \textbf{\textsc{Supervision}} & \textbf{\textsc{SentEval?}} & \textbf{\textsc{Base Model}} & \textbf{\textsc{Component}} & \textbf{\textsc{Average}} \\ \hline
        
         \citet{chen-etal-2022-generate} & Supervised (semi) & No & \texttt{t5} & \modelpill{} & 85.19 \\

         \citet{gao-etal-2021-simcse} & Unsupervised & Yes & \texttt{roberta-large} & \datapill{} & 83.76 \\
        
        \citet{ni-etal-2022-sentence} & Supervised & Yes & \texttt{t5} & \modelpill{} & 83.34 \\  
        
        \citet{case-augmented-positives} & Unsupervised & No & \texttt{roberta-large} & \datapill{} & 80.84 \\ 
        
        \citet{zhang-etal-2022-contrastive} & Unsupervised & Yes & \texttt{sbert-large} & \losspill{} & 80.69 \\  
        
        \citet{wang-lu-2022-differentiable} & Unsupervised & No & \texttt{bert-base} & \datapill{} & 80.61 \\

        \citet{liu-etal-2023-rankcse} & Unsupervised & Yes &  \texttt{roberta-large} & \losspill{} & 80.47 \\ 

        \citet{deng-etal-2023-clustering} & Unsupervised & Yes & \texttt{bert-large} & \datapill{} \losspill{} & 80.30 \\ 
        
        \citet{wu-etal-2022-infocse} & Unsupervised & Yes & \texttt{bert-large} & \losspill{} & 80.18 \\

        \citet{seonwoo-etal-2023-ranking} & Unsupervised & Yes & \texttt{bert-base} & \losspill{} & 80.07 \\ 
        
        \citet{wu-etal-2022-pcl} & Unsupervised & Yes & \texttt{bert-large} & \datapill{} & 79.94 \\  
        
        \citet{kim-etal-2021-self} & Unsupervised & Yes & \texttt{roberta-large} & \datapill{} & 79.76 \\  
        \citet{chen-etal-2023-alleviating} & Unsupervised & Yes & \texttt{bert-large} & \datapill{} & 79.69 \\
        
        \citet{wu-etal-2022-esimcse} & Unsupervised & Yes & \texttt{roberta-large} & \datapill{} & 79.45 \\  
        
        \citet{zhou-etal-2022-debiased} & Unsupervised & Yes & \texttt{roberta-large} & \datapill{} & 79.30 \\  
        
        \citet{wu-etal-2022-smoothed} & Unsupervised & No & \texttt{roberta-large} & \losspill{} & 79.21 \\  
        
        \citet{jiang-etal-2022-promptbert} & Unsupervised & No & \texttt{roberta-base} & \losspill{} & 79.15 \\  
        
        \citet{cao-etal-2022-exploring} & Unsupervised & Yes & \texttt{bert-large} & \datapill{} & 79.13 \\  
        
        \citet{zhang-etal-2022-virtual} & Unsupervised & No & \texttt{roberta-large} & \datapill{} & 79.04 \\ 
        
        \citet{mixing-negatives} & Unsupervised & Yes & \texttt{bert-large} & \datapill{} & 78.80 \\  
        
        \citet{min-etal-2021-locality-preserving} & Unsupervised & Yes & \texttt{bert-large} & - & 78.79 \\ 
        
        \citet{chuang-etal-2022-diffcse} & Unsupervised & Yes & \texttt{bert-base} & \losspill{} & 78.49 \\  
        
        \citet{jiang-etal-2022-improved} & Unsupervised & Yes & \texttt{bert-base} & \losspill{} & 78.49 \\
        
        \citet{chen-etal-2022-information} & Unsupervised & Yes & \texttt{roberta-large} & \losspill{} & 78.08 \\  
        
        \citet{wu-etal-2022-pcl} & Unsupervised & Yes & \texttt{roberta-base} & \datapill{} & 77.91 \\  
        
        \citet{cheng-2021} & Supervised & No & \texttt{roberta-large} & - & 77.47 \\  
        
        \citet{nishikawa-etal-2022-ease} & Unsupervised & No & \texttt{bert-base} & \datapill{} & 77.00 \\  
        
        \citet{reimers-gurevych-2019-sentence} & Supervised & Yes & \texttt{roberta-large} & \transformpill\losspill & 76.68 \\  
        
        \citet{liu-etal-2021-fast} & Unsupervised & No & \texttt{roberta-base} & \datapill{} & 76.40 \\ 

        \citet{wu-zhao-2022-sentence} & Unsupervised & No & \texttt{bert-base} & \losspill{} & 76.16 \\ 
        
        \citet{schick-schutze-2021-generating} & Unsupervised & No & \texttt{roberta-base} & \datapill{} & 75.20 \\
        
        \citet{klein-nabi-2022-scd} & Unsupervised & Yes & \texttt{bert-base} & \datapill{} & 74.19 \\  

        \citet{huang-etal-2021-whiteningbert-easy} & Unsupervised & No & \texttt{LaBSE} & \transformpill{} & 71.71 \\
        
        \citet{giorgi-etal-2021-declutr} & Unsupervised & Yes & \texttt{roberta-base} & \datapill{} & 69.99 \\  

        \citet{yang-etal-2021-universal} & Unsupervised & No & \texttt{bert-base} & \losspill{} & 67.22 \\
        
        \citet{zhang-etal-2020-unsupervised} & Unsupervised & Yes & \texttt{bert-base} & \losspill{} & 66.58 \\  
        
        \citet{li-etal-2020-sentence} & Unsupervised & No & \texttt{bert-base} & \datapill{} & 66.55 \\  \hline 
    \end{tabular}
    \caption{Comparison of methods. \textsc{supervision} indicates whether the method is supervised or unsupervised,  \textsc{SentEval} indicates whether the work benchmarks against SentEval \cite{conneau-kiela-2018-senteval}, \textsc{component} indicates the component from \Cref{fig:components} that the work targets, and \textsc{average} is the average score on STS.}
    \label{tab: results}
\end{table*}
\subsection{Natural Language Inference}

% - Learning NLI requires understanding the semantics

% - Learning NLI well, leads to better sentence representations

%     - Using Bidirectional sentence representation - SentenceBERT, Dual View DistilBERT
    
%     - Using Generative Transformers - Sentence T5

%     - Using the same technique 

Natural Language Inference (NLI) is the process of determining the logical relationship between a premise (an assumed true sentence) and a hypothesis (a possibly true sentence). The objective of NLI is to determine whether the hypothesis can be logically inferred from the premise (entailment), contradicts the premise (contradiction), or is neutral with respect to it \cite{dagan2013recognizing}. NLI serves as a proxy for evaluating natural language understanding. According to \citet{conneau-etal-2017-supervised}, learning sentence representations using NLI data can be effectively transferred to other NLP tasks, demonstrating the generality of this approach.

 \citet{reimers-gurevych-2019-sentence} and subsequent works mainly rely on learning sentence representations using NLI data. There are two noteworthy components to enable this. First, processing inputs individually without promoting interaction between words; second, using an encoder like BERT  as its backbone model. The first component is computationally efficient but has been found to result in poorer performance compared to methods that promote interaction between words \cite{reimers-gurevych-2019-sentence}. This lack of interaction can limit the network's ability to capture the nuances of language, and may result in less accurate sentence embeddings. In order to solve this, efforts such as the work from \citet{cheng-2021},  incorporated word-level interaction features into the sentence embedding while maintaining the efficiency of Siamese-BERT networks. Their approach makes use of ideas from knowledge distillation \cite{hinton2015distilling}: using the rich knowledge in pretrained cross-encoders to improve the performance of Siamese-BERT.

Meanwhile, with the raise of generative models which have a myriad of capabilities has  lead researchers to explore whether they can serve as better backbone models for sentence representations \cite{ni-etal-2022-sentence} compared to encoder-only models like BERT. They consider three methods to obtain sentence representations from a pretrained T5 model: the representation of the first token of the encoder,  the representation of the first generated token of the decoder, or the mean of the representations from the encoder. They found that such models trained on NLI are performant, showing the utility of generative models for obtaining sentence representations.

\subsection{Generating Data}
Acquiring supervised data to train sentence representations is a difficult task. However, in recent years, pre-trained models have emerged as a potential solution for generating training data. Furthermore, pre-trained models can serve as weak labelers to create ``silver data''.

Cross-encoders that are pretrained on NLI data can be used to obtain silver data. In order to do this, \citet{thakur-etal-2021-augmented} suggest Augmented-SBERT. Their approach involves using different strategies to mine sentence pairs, followed by labeling them using a cross-encoder to create silver data. The silver data is then combined with the human-labelled training dataset, and a Siamese-BERT network is trained. However, this method requires mining appropriate sentence pairs first. 

Rather than relying solely on obtaining supervised data, researchers are exploring the use of generative language models to create large amounts of synthetic training data for sentence encoders. This approach has the potential to produce high-quality training data at scale, addressing some of the challenges associated with supervised data acquisition. For instance, \citet{chen-etal-2022-generate} demonstrate the use of a T5 model trained to generate entailment or contradiction pairs for a given sentence. However, this method still needs to provision a sentence to generate the entailment/contradiction pairs.

DINO, introduced by \citet{schick-schutze-2021-generating}, automates the generation of NLI data instructions using GPT2-XL. This approach eliminates the need for providing a sentence to generate entailment or contradiction pairs. Models trained on the resulting STS-Dino dataset outperform strong baselines on multiple semantic textual similarity datasets.

\section{Unsupervised Sentence Representations}
\label{sec:unsupervised}
Unlike supervised methods, unsupervised learning techniques do not rely on explicit positive and negative examples but instead employ alternative techniques to mine them. Hence this has garnered significant attention in recent years. Additionally, they may also modify the learning objectives.

\subsection{Better Positives} 
\label{subsec:better-positives}
 Contrastive learning techniques optimize sentence representations by contrasting semantically similar examples against dissimilar ones (c.f \Cref{subsec:sentence-representation-components}). A simple way to obtain a semantically similar example is to make minimal changes to it. In contrast to images, where simple transformations such as rotation, clipping, and color distortion can generate semantically similar examples, deleting or replacing a random word in a sentence can drastically change its meaning \cite{schlegel2021semantics}. Therefore, it is crucial to carefully select positive and negative examples for contrastive learning in NLP.

\subsubsection{Surface Level}
To create a sentence that carries the same meaning as another, one can modify the words or characters in the text. 
Recent research \cite{case-augmented-positives,liu-etal-2021-fast,wu-etal-2022-esimcse} suggests certain transformations that preserve the semantic meaning. \citet{case-augmented-positives} propose randomly flipping the case of some tokens, while \citet{liu-etal-2021-fast} mask spans of tokens to get positive instances, and \citet{wu-etal-2022-esimcse} suggest to repeat certain words or subwords.  Besides generating positive instances, these transformations help in fixing certain biases in representations generated by transformers. For example, \citet{jiang-etal-2022-promptbert} found that avoiding high-frequency tokens can result in better sentence representations, and transformations that mask them out while learning sentence representations can improve its quality.

However, altering the surface characteristics of sentences can lead to models relying on shortcuts rather than learning semantics \cite{du-etal-2021-towards}. To address this issue, \citet{wu-etal-2022-pcl} propose the use of multiple augmentation strategies rather than a single transformation. They use shuffling, repeating, and dropping words as transformation strategies to improve model robustness. Additionally, they implement mechanisms to enhance learning from multiple positive examples.

\subsubsection{Model Level}
\abhinav{Minor modifications to the words or the structure of a sentence can still result in big changes in semantics in language processing. However, researchers have explored another method, where such small modifications can be made in the representation space} by leveraging the distinctive characteristics of the backbone model utilized in contrastive learning. These characteristics might be architectural choices, or using representations from certain components of the model. 

One such approach uses Dropout -- a regularization technique used in deep learning to prevent overfitting of a model. During training, some neurons in the layer are randomly deactivated, resulting in slightly different representations when the same training instance is passed through the model multiple times. These different representations can be used as positive examples for learning. Recent studies such as \citet{gao-etal-2021-simcse} have demonstrated the effectiveness of dropout as an augmentation strategy. Several other works have also incorporated this technique and improved upon it: promoting decorrelation between different dimensions \cite{klein-nabi-2022-scd} and adding dropout in the transformation arsenal \cite{wu-etal-2022-pcl, wu-etal-2022-esimcse}.

On the other hand, special components can be trained to generate semantically similar representations. One example is the use of prefix modules~\cite{li-liang-2021-prefix}, which are small, trainable modules added to a pretrained language model. \citet{wang-lu-2022-differentiable} attach two prefix modules to the Siamese BERT network (c.f \Cref{sec:background}) -- one each for the two branches -- and train them on NLI data. This enables the prefix modules to understand the nuances of the difference between representations. They show that representations from the two modules for the same sentence can then be  used as positives. 

\subsubsection{Representation Level}
Examining the latent representation of sentences generated by a model yields a valuable benefit. In this scenario, one can discover positive examples by exploring the representation space. These approaches offer the distinct advantage of obviating the need for any data augmentation.

Although BERT's \texttt{[CLS]} representation is commonly used as a sentence representation, it has been shown to be ineffective~\cite{reimers-gurevych-2019-sentence}. In fact, \citet{kim-etal-2021-self} demonstrated that the various layers of BERT have differing levels of performance on the STS dataset. To address this issue, they propose reusing the intermediate BERT representations as positive examples. In contrast, \citet{zhang-etal-2022-virtual} perform augmentation by identifying the $k$-nearest neighbors of a sentence representation. 

\subsubsection{Alternative Methods}
Researchers have explored various other methods for obtaining positive samples for unsupervised sentence representations. One option is weak supervision: using spans from the same document \citep{giorgi-etal-2021-declutr}, employing related entities \citep{nishikawa-etal-2022-ease}, and utilizing tweets and retweets-with-quotes \citep{di-giovanni-brambilla-2021-exploiting}. On the other hand, dialogue turns can be used as semantically related pairs of text for learning sentence representations \cite{zhou-etal-2022-learning}.

Other approaches use the capability of large language models to perform tasks based on instructions---a technique called ``prompting''. Researchers have used prompts to obtain better sentence representations, as demonstrated in studies such as \citet{jiang-etal-2022-promptbert}, which employs the \emph{``[X] means [MASK]''} prompt to extract sentence representations from the representation of the \emph{``[MASK]''} token in a sentence. Another study by \cite{zeng-etal-2022-contrastive} combines prompt-derived sentence representations with contrastive learning to improve the quality of the representations.

% What are some key takeaways ?
% How do the different methods compare against each other  - Which one is more efficient? Which one makes more or less assumptions 
% What are some trends that you see and what is your take on this 
% What do you think the future will be

\subsection{Alternative Loss and Objectives}
\label{subsec:alternative-loss-objectives}
In \Cref{sec:background} we discussed Contrastive loss, which is widely used in machine learning. However, this loss suffers from several limitations: for instance it only considers binary relationships between instances and lacks a mechanism to incorporate \textit{hard-negatives} (negatives that are difficult to distinguish from positive examples). To overcome these drawbacks, researchers have explored various strategies:

\paragraph{Supplementary Losses:} Used in addition to contrastive losses, these include: \emph{(1)}  hinge loss \citep{jiang-etal-2022-improved}, which enhances discrimination between positive and negative pairs; \emph{(2)} losses for reconstructing the original sentence from its representation to better capture sentence semantics \citep{wu-etal-2022-infocse} ; \emph{(3)} a loss to identify masked words and improve sensitivity to meaningless semantic transformations \citep{chuang-etal-2022-diffcse}; and \emph{(4)} a loss to minimize redundant information in transformations by minimizing entropy \citep{chen-etal-2022-information} \abhinav{(5) Ranking based losses to ensure that all negatives are not treated equally -- some negatives are closer to the query compared to others \cite{seonwoo-etal-2023-ranking, liu-etal-2023-rankcse}}

\paragraph{Modified Contrastive Loss:}  \citet{wu-etal-2022-smoothed} proposed an additional term  that incorporates random noise from a Gaussian distribution as negative instances. Also, \citet{zhang-etal-2022-contrastive} introduced two losses, angular loss and margin-based triplet loss, to address the intricacies of similarity between pairs of examples.

\paragraph{Different Loss:} Moving away from contrastive loss. \abhinav{Disadvantages of contrastive representations include not considering the relevance of different parts of the sentence in the entire  representation, and assuming that sentence representations lie in the Euclidean space. \citet{zhang-etal-2020-unsupervised} address the first by maximizing the mutual information between a local context and the entire sentence. \citet{min-etal-2021-locality-preserving} address the second by identifying  an alternative sub-manifold within the sentence representation space.} Other objectives to learn sentence representations include disentangling the syntax and semantics from the representation \cite{huang-etal-2021-disentangling}, generating important phrases from sentences instead of using contrastive learning \cite{wu-zhao-2022-sentence}, or using sentence representation as a strong inductive bias to perform Masked Language Modeling \cite{yang-etal-2021-universal}.

\subsection{Better Negative Sampling}
\label{subsec:better-negative-sampling}
The efficacy of contrastive learning hinges on the quality of negative samples used during training. While most methods prioritize selecting positive samples that bear similarity to the query text, it is equally crucial to include hard negatives that are dissimilar to the query text and pose a challenge for the model to classify. Failure to do so leads to a gradual diminution of the loss gradients, impeding the learning of useful representations \cite{mixing-negatives}. Additionally, using an adequate number of negative samples is also imperative for effective learning \cite{cao-etal-2022-exploring}.

Given the importance of incorporating hard negatives,  several innovative strategies have emerged. Researchers have found that mixed-negatives---a combination of representations of a positive and a randomly chosen negative---serve as an excellent hard negative representation \citep{mixing-negatives}. Similarly, \citet{zhou-etal-2022-debiased}  leveraged noise from a uniform Gaussian distribution as negatives to foster uniformity in the learned representation space---a metric to assess learned sentence representation. Recently,  \abhinav{In contrast to the approach taken by \citet{kim-etal-2021-self}, \cite{chen-etal-2023-alleviating} employ representations from various layers as negatives, recognizing that similarities across these layers render them less discriminative. This contemporary approach shows enhanced performance on the STS benchmark and subsequent tasks. However, it's important to note that perceptions of what constitutes 'positive' or `negative' in the literature are constantly evolving.}

\abhinav{False negatives are instances where certain negatives exhibit a higher similarity to the anchor sentence compared to other negatives, yet maintain a lower similarity than the positives. Properly identifying and integrating measures to address these false negatives is crucial for enhancing sentence representation learning. \cite{deng-etal-2023-clustering} tackle this by clustering the remaining N-1 sentences in a batch. Sentences within the same cluster are designated as false negatives. To manage this scenario effectively, they employ a Bidirectional Margin Loss. This approach ensures that false negatives are not excessively distanced from the anchor sentence, thereby improving the overall quality of the sentence representation. }

\subsection{Post-Processing}
\label{sec:make-plms-better}
 \citet{ethayarajh-2019-contextual} suggest that the out-of-the-box representations from LLMs are not effective sentence representations. Consequently, several efforts have addressed this issue.

\citet{almarwani-etal-2019-efficient} utilize the Discrete Cosine Transform, a widely used technique in signal processing, to condense word vectors into fixed-length vectors. This approach has demonstrated its effectiveness in capturing both syntax and semantics. Similarly, \citet{li-etal-2020-sentence} employ normalizing flows to convert BERT's token representations into a Gaussian distribution, while \citet{huang-etal-2021-whiteningbert-easy} propose a simpler `whitening' technique that enhances out-of-the-box sentence representations from LLMs by transforming the mean and covariance matrix of the sentence vectors. \abhinav{These post processing techniques have only been tested on BERT based models so far, and their generalization to newer models has not been answered.}

\section{Other Approaches}
\label{sec:other}

\paragraph{Multimodal:}
Human experiences are complex and involve multiple sensory modalities. Thus, it is beneficial to incorporate multiple modalities in learning sentence representations. Researchers have explored different approaches to use images for this purpose: using contrastive loss that utilizes both images and text \citep{zhang-etal-2022-mcse}; optimizing a loss each for visual and textual representation \citep{jian-etal-2022-non}; grounding text into image  \citep{bordes-etal-2019-incorporating}. Other modalities like audio and video are yet to be incorporated. \abhinav{Given that obtaining supervised data with just one modality is already hard, obtaining the same for multiple modalities will be even more challenging.}

\paragraph{Computer Vision Inspired:}
Momentum encoder, introduced by \citet{momentum-encoder-he}, improves training stability. It utilizes a queue of representations from previous batches as negatives for the current batch, decoupling batch size from the learning process. Several studies have integrated momentum encoder into sentence representation learning, leading to enhanced performance~\cite{cao-etal-2022-exploring, wu-etal-2022-pcl, wu-etal-2022-esimcse, tan-etal-2022-sentence}. \abhinav{This might require additional memory in the GPU which is challenging when training large NLP models.}

Another popular technique, Bootstrap Your Own Latent (BYOL)~\cite{byol-grill}, is a self-supervised learning method that dispenses with negative samples. It trains a neural network to predict a set of `target' representations from an input data point, given an `online' representation of the same data point. BYOL employs a contrastive loss function to encourage similarity between the online and target representations. An advantage of BYOL is the elimination of the need for negative samples; instead, it uses augmented versions of the same data point as positive samples. This method has been effectively applied to natural language processing by \citet{zhang-etal-2021-bootstrapped} \abhinav{It implicitly assumes that obtaining an augmented sentence is easy -- which might not be the case, as we have seen in the previous sections}.

\section{Trends \& Analysis}
\label{sec:analysis}

\paragraph{Limited advantages of supervision: } \Cref{tab: results} summarizes all the results. Surprisingly, a simple dropout-based data augmentation technique \cite{gao-etal-2021-simcse} demonstrates superior performance compared to most other methods, including those using T5, which is trained on billions of tokens \cite{ni-etal-2022-sentence}. \abhinav{Note that T5 is trained on a token generation objective that might not be suitable for obtaining better sentence representations. Besides the model, using an appropriate unsupervised task might be important for better representations.} 

\paragraph{Downplaying downstream task evaluation: } The neglect of evaluating sentence representations in downstream tasks, as exemplified in \Cref{tab: results}, is noticeable. With LLMs demonstrating remarkable zero-shot performance across various tasks, the utility of sentence representations for tasks beyond semantic similarity and retrieval seems to dwindle. Nevertheless, recent research shows how sentence representations can enhance few-shot text classification performance \cite{tunstall2022efficient}. \abhinav{Future sentence representations should consider the utility of representations in enhancing few-shot text classification.}

\paragraph{Data-centric innovations: } Most innovations in this field focus on improving the \textbf{data} aspect, including obtaining better positives or negatives, and generating data using large language models \cite{schick-schutze-2021-generating, chen-etal-2022-generate}. While generative models like T5 can boost performance, other LLMs like ChatGPT can bring additional benefits because of their scale.

\paragraph{Keeping up with LLMs:} We have identified several noteworthy endeavors using massive language models with billions of parameters for sentence representations. SGPT \cite{muennighoff2022sgpt} has successfully trained an open-source GPT decoder-only model on the SNLI and MNLI datasets, surpassing OpenAI's 175B parameter model. Additionally, GTR \cite{ni-etal-2022-large} examined scaling laws, revealing larger T5 models have better performance. Nonetheless, recent developments such as GTE \cite{li2023general} and BGE \cite{bge_embedding} highlight that a collection of high-quality datasets for contrastive training can yield significantly better results compared to just using bigger models.

\section{Challenges}
\label{sec:challenges}

\paragraph{Practical Applications and the rise of Tools:} Sentence representations are commonly employed for sentence retrieval in practical applications, as evidenced by the increasing number of benchmarks \citep{thakur2021beir}.  However, their utility extends beyond retrieval, as demonstrated by recent work \cite{schuster-etal-2022-stretching}, which  leverages sentence representations for identifying documents that share a similar stance on a topic and for isolating documents that diverge from the consensus.

The increasing use of sentence representations in practical applications such as retrieval and \abhinav{providing an appropriate context to generative language models to rely on has lead to the rise of tools known as vector databases. These tools enable storing vectors as indices and include algorithms for fast retrieval of similar vectors. Popular options such as Pinecone\footnote{https://www.pinecone.io/} and Milvus\footnote{https://milvus.io/} also offer services for cloud hosting and resilience.}  These vector databases can be integrated with other frameworks such as LangChain, that facilitate the development of LLM applications.

\paragraph{Adapting to Different Domains:} Research has shown that sentence representations learned in one domain may not accurately capture the semantic meaning of sentences in another domain \cite{jiang-etal-2022-improved, thakur-etal-2021-augmented}. Some solutions have been proposed in the literature, such as generating queries using a pretrained T5 model on a paragraph from the target domain, or using a pretrained cross-encoder to label the query and paragraph, or using a denoising objective \cite{wang-etal-2021-tsdae-using}. Nonetheless, training models that work well across domains remains challenging. 

\paragraph{Cross-lingual Sentence Representations:} Creating sentence representations that can be used across languages, especially those with limited annotated data, poses a significant challenge \cite{miracl}. New solutions for cross-lingual retrieval are being developed and deployed for real-world use cases.\footnote{https://txt.cohere.com/multilingual/}  Many scholarly works \cite{nishikawa-etal-2022-ease, feng-etal-2022-language, wieting-etal-2020-bilingual} have addressed cross-lingual sentence representation learning in recent times, but they require aligned data between languages, which is hard to obtain.

\paragraph{Universality of Sentence Representations: }
The original purpose of sentence representations was to serve as a versatile tool for various NLP tasks. One prominent effort to evaluate the universality of sentence representations was the SentEval task \cite{conneau-kiela-2018-senteval}, which tested the representations' performance on text classification, natural language inference, and semantic text similarity tasks. However, many recent works on sentence representation tend to emphasize their effectiveness on semantic text similarity datasets (\Cref{tab: results}). This shift raises questions about the universal nature of these representations---are sentence representations useful only for retrieval, or do they indeed have other applications? Such questions are put back into spotlight by recent benchmarks such as MTEB \cite{muennighoff2022mteb}.

\section{Conclusions}
This survey offers an overview of sentence representations, presenting a taxonomy of methods. While major innovations focused on obtaining better quality data for contrastive learning, modern advances in generative technologies can accelerate the automatic generation of supervised data at low cost. Although LLMs play a crucial role in informing the advancement of sentence representations, further enhancements in sentence representation learning are necessary to personalize current LLMs to achieve tailored results. We highlighted that better multilingual and multi-domain sentence representations are needed, now that LLMs are being deployed in different domains at a rapid pace. We hope that this survey can accelerate advances in sentence representation learning.

\section{Limitations}
While we have made an effort to encompass a comprehensive range of literature on sentence representations, it is possible that certain papers may have been inadvertently excluded from our literature review. Additionally, we acknowledge that our approach assumes the majority of methods primarily focus on sentences or a limited number of tokens, typically within a few hundred. However, it is important to note that representation learning for documents or longer contexts---an active area of research---utilizes similar techniques. This survey does not cover those specific areas, which may warrant further attention.

% \section*{Acknowledgements}

\bibliography{custom}
\bibliographystyle{acl_natbib}

\end{document}